\documentclass[sigconf]{acmart}
\AtBeginDocument{%
  \providecommand\BibTeX{{%
    \normalfont B\kern-0.5em{\scshape i\kern-0.25em b}\kern-0.8em\TeX}}}

\setcopyright{acmcopyright}
\copyrightyear{2024}
\acmYear{2024}
\acmDOI{XXXXXXX.XXXXXXX}

\acmConference[DAC'24]{Make sure to enter the correct
  conference title from your rights confirmation email}{June 23--27,
  2024}{San Francisco, CA}
\acmPrice{15.00}
\acmISBN{978-1-4503-XXXX-X/18/06}

\usepackage{algorithm2e}
\usepackage{mathtools}
\RestyleAlgo{ruled}
\SetKwComment{Comment}{/* }{ */}
\usepackage{xcolor}
\usepackage{adjustbox}

\definecolor{myBlue}{rgb}{0.0,0.0,1.0}

\begin{document}

\title{OPAL : \underline{O}utlier-\underline{P}reserved Microscaling Quantization \underline{A}ccelerator for Generative \underline{L}arge Language Models}
%%\title{OPAL : \underline{O}utlier-\underline{P}reserved \underline{A}daptive Microscaling Quantization for \underline{L}arge Language Model Generation Tasks}


\author{Jahyun Koo}%
\authornote{J. Koo and D. Park are equally contributed authors.}
\affiliation{%
  \institution{DGIST}
  \country{South Korea}
}
\email{jhkoo@dgist.ac.kr}

\author{Dahoon Park}%
\authornotemark[1]
\affiliation{%
  \institution{Korea University}
  \country{South Korea}
}
\email{manyteacher93@korea.ac.kr}
\author{Sangwoo Jung}
\affiliation{%
  \institution{DGIST}
  \country{South Korea}
}
\email{jsangwoo123@dgist.ac.kr}

\author{Jaeha Kung}%
\authornote{J. Kung is the corresponding author (jhkung@korea.ac.kr).}
\affiliation{%
  \institution{Korea University}
  \country{South Korea}
}
\email{jhkung@korea.ac.kr}

\renewcommand{\shortauthors}{J. Doe et al.}

\begin{abstract}
% Large language model (LLM) size has been increasing 10$\times$ every year for the last few years.
% This trend naturally demands a higher memory footprint and a wider memory bandwidth for servicing LLMs.
% To overcome such burden on the memory size and bandwidth, aggressive weight quantization on LLMs has been recently studied, while lacking research on quantizing activations. 
To overcome the burden on the memory size and bandwidth due to ever-increasing size of large language models (LLMs), aggressive weight quantization has been recently studied, while lacking research on quantizing activations. 
In this paper, we present a hardware-software co-design method that results in an energy-efficient LLM accelerator, named OPAL, for generation tasks.
First of all, a novel activation quantization method that leverages the microscaling data format while preserving several outliers per sub-tensor block (e.g., four out of 128 elements) is proposed.
Second, on top of preserving outliers, mixed precision is utilized that sets 5-bit for inputs to sensitive layers in the decoder block of an LLM, while keeping inputs to less sensitive layers to 3-bit.
Finally, we present the OPAL hardware architecture that consists of FP units for handling outliers and vectorized INT multipliers for dominant non-outlier related operations.
In addition, OPAL uses log2-based approximation on softmax operations that only requires shift and subtraction to maximize power efficiency.
As a result, we are able to improve the energy efficiency by 1.6$\sim$2.2$\times$, and reduce the area by 2.4$\sim$3.1$\times$ with negligible accuracy loss, i.e., <1 perplexity increase.

% We present a energy efficiency hardware accelerator, called outlier-preserved microscaling quantization accelerator for generative large language models (OPAL), by introducing novel algorithm-hardware co-optimization technique. 
% Our hardware has improved quantization performance by effectively preserving the activation outlier, thereby increasing the computational efficiency of LLMs. 
% We then provide a log2-based approximation for hardware-intensive softmax operations, replacing operations that previously required division and multiplication with subtraction and shift. 
% Finally, we show that using a reconfigurable vector multiplier to enforce mixed bit-precision for LLM can achieve 2.1$\times$/1.6$\times$  computational efficiency and reduce 3.7$\times$/2.83$\times$ area size instead of only less than 0.5 and 1 ppl increase over the conventional vector multiplier.
\end{abstract}

\maketitle

\section{Introduction}

Large-scale language models (LLMs) have been a game changer that work well in various language-related tasks, including translation~\cite{bert} and text-to-image diffusion~\cite{imagen}, to name a few. 
However, due to the massive model size, running a generation task using LLMs is economically expensive and consumes substantial energy even at servicing inference. 
For instance, to run the Llama2-70B model~\cite{llama}, it needs to store 140GB of memory using FP16, and it requires at least 140 GFLOPs to generate only one token. 
In addition, it takes at least two A100 80GB GPUs to perform inference that costs more than \$200K.
% $200k는 어디서 구한 수치인지? # A100 2대의 비용으로 지정해두었습니다.
As the number of parameters in an LLM keeps growing, e.g., GPT3-175B, service costs are expected to increase dramatically. 
To mitigate this challenge in deploying LLMs, various compression techniques on LLMs have been actively studied to fit the model into single-GPU memory capacity~\cite{gptq,owq,llm_prune}.

\begin{figure}[t]
  \includegraphics[width=0.8\linewidth]{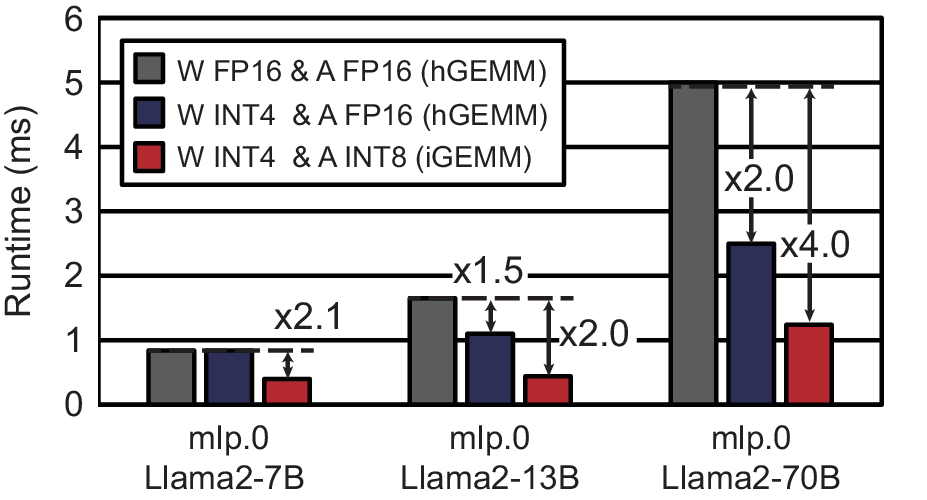}
  \caption{The comparison of single-batch latency in running the Llama2's first FFN layer (\textit{mlp.0}) at various sizes and bit-widths using CUTLASS~\cite{cutlass}. The `\texttt{hGEMM}' uses FP16 computing units while `\texttt{iGEMM}' uses INT8 computing units.}
    \label{intro}\vspace{-5mm}
\end{figure}

Among various compression techniques, quantization has been prevailed upon others due to its effectiveness in both performance boost and memory reduction.
Especially, due to high training time of LLMs, post-training quantization (PTQ) has become the de facto standard in quantizing LLMs~\cite{gptq,owq,zeroquant,llmint8}.
% either activations and/or weights of 
The common observation from the prior work on quantizing LLMs is that some input channels in activations consistently produce large values, i.e., \textit{outliers}, allocating few bits to non-outliers resulting in high quantization errors.
Therefore, several studies have identified those activation outliers during the \textit{weight-only quantization}~\cite{gptq,owq}.
Quantizing weights is extremely effective since generation tasks are memory-bounded due to small arithmetic intensity (FLOPs/Byte).
However, when aggressive weight quantization is applied, e.g., 4-bit, the workload moves toward the compute-bounded region necessitating the need for activation quantization as well (\textbf{motivation 1}).
% so that more area-efficient thus more arithmetic units can be utilized.
Fig.~\ref{intro} shows that quantizing weights of (memory-bounded) Llama-13B and Llama-70B models to INT4, while computing with FP16 units (\texttt{hGEMM} in CUTLASS), reduces the runtime by 1.5$\times$ and 2.0$\times$, respectively.
This is mainly due to the higher core utilization according to the increased arithmetic intensity thanks to the quantization.

There is another research direction trying to quantize both activations and weights~\cite{zeroquant,llmint8}.
The prior work successfully quantized both tensors to INT8 with little accuracy loss (but tested a model with up to 13B parameters).
With 8-bit activations, we now can utilize INT8 computing units within a GPU pushing the computational roof higher (\texttt{iGEMM} in CUTLASS).
%If the batch prompting or paged attention is utilized, we can further quantize activations to less than 8-bit to achieve higher throughput. 
We tested Llama2 models at various sizes while quantizing the weights to INT4 and activations to INT8.
As a result, it was possible to achieve 2.0$\sim$4.0$\times$ speed-up for all tested models thanks to the higher number of available computing units.
However, quantizing activations on the fly requires min/max extraction and dividers to account for scale factors, which incur significant hardware overhead (\textbf{motivation 2}).
% min/max extraction도 on the fly로 필요한 게 맞는지?

With these motivations in mind, we present a hardware accelerator for LLM-based generation tasks, dubbed OPAL (\textbf{\underline{O}}utlier-\textbf{\underline{P}}reserved microscaling quantization \textbf{\underline{A}}ccelerator for \textbf{\underline{L}}LMs), which is based on a novel algorithm-hardware co-design method.
The OPAL utilizes a microscaling data format (MX format~\cite{microscaling}) that allows quantizing activations to 3/5-bit or 4/7-bit with the support of preserving several outliers within a pre-defined block.
%called \textbf{O}utlier-\textbf{P}reserved microscaling quantization \textbf{A}ccelerator for \textbf{L}arge language model generation task (\textbf{OPAL}), 
The microscaling data format allows us to replace dividers to simple shifters to reflect scale factors, which is more suitable for dynamic quantization.
%It supports dynamic post training quantization without a divider using outlier-preserved microscaling floating point format, then accelerates a reconfigurable PE that supports this format is xx more computationally efficient. 
In addition, OPAL hardware is equipped with lightweight softmax units and reconfigurable PEs that improve the energy efficiency by 2.41/3.18$\times$ compared to the bfloat16 baseline.
Our main contributions can be summarized as follows:
\begin{itemize}
    \item \textbf{\textit{Shift-based Dynamic Quantizer}}: We propose an \textit{outlier-preserved microscaling integer format} (MX-OPAL) to implement shift-based dynamic quantizers. 
    Since activations in LLMs have outliers, na\"ive MXINT8 leads to significant accuracy degradation. 
    Thus, in this work, we keep a small number of outliers, i.e., four largest absolute values per block, in bfloat16 while quantizing non-outliers in 3/5-bit or 4/7-bit.
    %Since activation in the LLM model has out of range value, naive MXINT quantization leads to a significant performance decrease. Therefore, we propose an outlier-preserved MXINT quantization, which stores the two largest absolute values in each block for activation as BF16 and quantizes the rest in blocks. Unlike traditional min-max quantization, which requires divider for quantization, our outlier-preserved MXINT quantization replaces divider with shift operation, reducing power consumption by a factor of x without sacrificing performance.
    \item \textbf{\textit{Hardware-friendly Softmax Approximation}}: We present log2-based approximation on softmax functions in an LLM to perform `\textit{softmax}(Q$\cdot$K\textsuperscript{T})$\cdot$V' with only shift-and-subtractions.
    The approximated softmax only increases the perplexity by <0.4 PPL on WikText-2 while saving 1.56$\times$ power consumption of conventional softmax.
    % We applied log2-based approximate softmax and the shifted accumulation attn\_output module to remove unnecessary dividers and multipliers. The approximate softmax resulted in a power saving of more than 2x on the WikiText task with a PPL reduction of less than 0.4 compared to traditional softmax.
     \item \textbf{\textit{Hardware Accelerator for LLM-based Generation}}: Finally, we present an energy-efficient hardware accelerator for generative LLM, named OPAL. 
     The OPAL enjoys power efficiency of shift-based quantizers and softmax units that are tested on recent LLMs.
     The OPAL improves the energy efficiency by 1.6$\times$/2.2$\times$ under PPL increase of 0.33/0.62 compared to the state-of-the-art weight-only quantized models.
     %We also report implementation results of OPAL designs by mapping various Large Language models (OPT, Llama2), including the recent Llama2 model for the first time, increasing the energy-efficiency by 2.09×/3.67x under PPL drop 0.5/1 compared state-of-the-arts weight-only quantization models.
    
\end{itemize}\vspace{-4mm}
\section{Background}
\subsection{Quantization for LLMs}

Weight-only quantization is the most popular approach to compress LLMs, since the weights occupy most of the memory space in LLM-based generation tasks.
%representing weights as a low-precision integer format to reduce the power consumption of data interconnects and storage space. 
OPTQ (also known as GPTQ~\cite{gptq}) looks at Hessian matrices to find sensitive input channels and apply quantization in order of sensitivity to achieve 3-/4-bit weights.
%is based on the Optical Brain Compression method, which uses Hessian matrices to compress weights to 3-bit or 4-bit into layer-wise post-training quantization. 
On top of OPTQ, OWQ~\cite{owq} applies mixed-precision quantization that keeps weights in FP16 at highly sensitive input channels, while others in 3-/4-bit.
% OWQ is a mixed-quantization algorithm modified from OPTQ that uses slightly more memory than OPTQ by storing significant weights to high bit (FP16) due to its multiplied to activation’s colossal outlier values~\cite{owq}. 
As a result, OWQ shows that 3.01-bit weights are enough to provide comparable LLM accuracy to the 4-bit OPTQ model (i.e., 25\% memory reduction).
However, since the prior work on \textit{weight-only quantization} performs matrix multiplication (matmul) in FP16, the computing energy is not reduced compared to the original FP16 model.
% AWQ에 대한 내용 일단 제외

Quantizing both inputs and weights to INT8~\cite{zeroquant,llmint8} reduces the inference energy by using more power-efficient INT8 units and fetching 2$\times$ less data from DRAM.
% Weight and activation quantification reduces the power used for LLM inference, considering computational efficiency as well as memory footprint.
ZeroQuant \cite{zeroquant} is a fine-grained compression method utilizing group-wise weight quantization and token-level activation quantization.
Unlike the weights, activations are dynamically quantized, i.e., extracting min/max range per token, to minimize quantization error since activations are input-specific.
In LLM.int8()~\cite{llmint8}, vector-wise normalization and outlier decomposition are used to maintain the LLM accuracy while quantizing most of the values to INT8 for both tensors.
%the distribution of activation is input-specific, so it improves performance by quantizing dynamically when input activation is calculated. 
However, dynamic quantization consumes too much power since floating-point (FP) divisions are required to convert output values to integer values with no accuracy loss. 
% SmoothQuant에 대한 내용 일단 제외

%\textbf{Quantization} is a widely used methodology that reduces memory usage by mapping the continuous high bit to the discrete low bit.  Quantization has the advantage of increasing computational efficiency and reducing memory usage but has the problem of lowering output quality. Quantization Aware Training (QAT) and Post Training Quantization (PTQ) are typical ways to address these issues. QAT trains with quantization noise to improve performance, which is challenging due to the massive cost of LLM training. On the contrary, PTQ is more frequently used for LLM quantization because it does not require training.

%OPTQ focuses on reducing the size of the LLM parameters and is one of the PTQs that only quantize weight, using a Hessian-based metric of layer-wise quantization errors, based on Optical Brain Compression \cite{}. OPTQ showed similar performance to the existing fp16 model at 4 bit, but there is a disadvantage of poor accuracy at 3 bit. OWQ modified OPTQ to store FP16 for some of the weights operated by the activation outlier, which, using only 3.01 bits, showed a similar performance to OPTQ's 4 bits. The above two approaches are great for reducing weight and solving the memory bottleneck. However, they have the disadvantage of still using FP16 for computation, which could be more efficient. We aim to reduce the computational bound by proposing an effective activation quantization method.

\begin{figure}[t]
  \includegraphics[width=\linewidth]{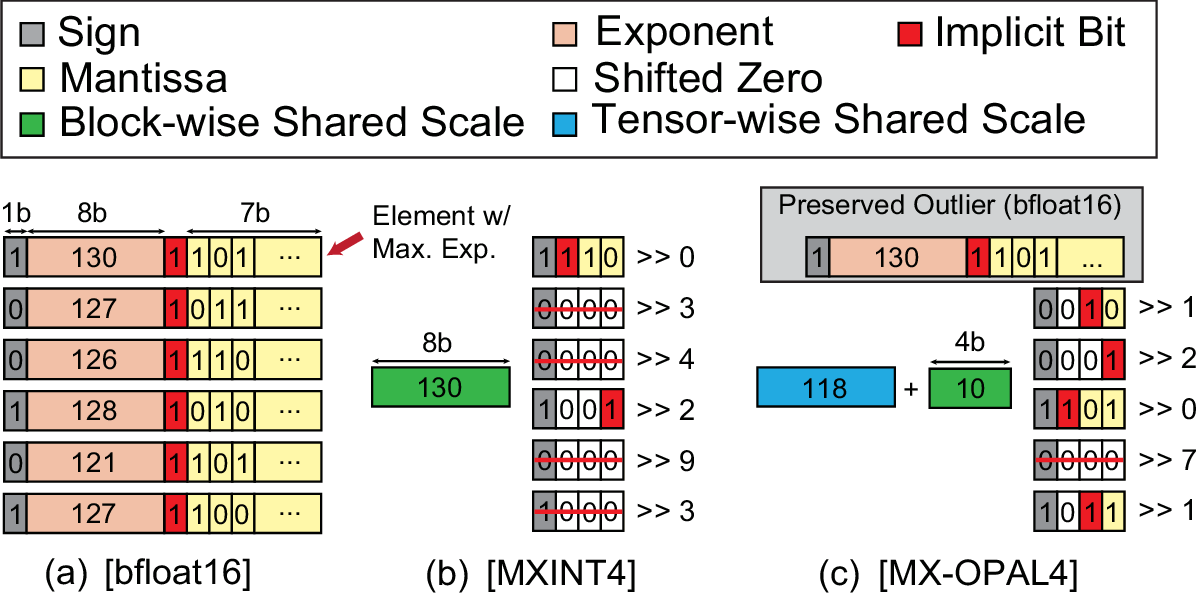}\vspace{-2mm}
  \caption{Various data formats: (a) bfloat16, (b) original MXINT4~\cite{microscaling}, and (c) the proposed outlier-preserved MXINT4 format (i.e., MX-OPAL4).
}
\label{BFP_format}\vspace{-4mm}
\end{figure}

\begin{figure}[t]
  \includegraphics[width=0.85\linewidth]{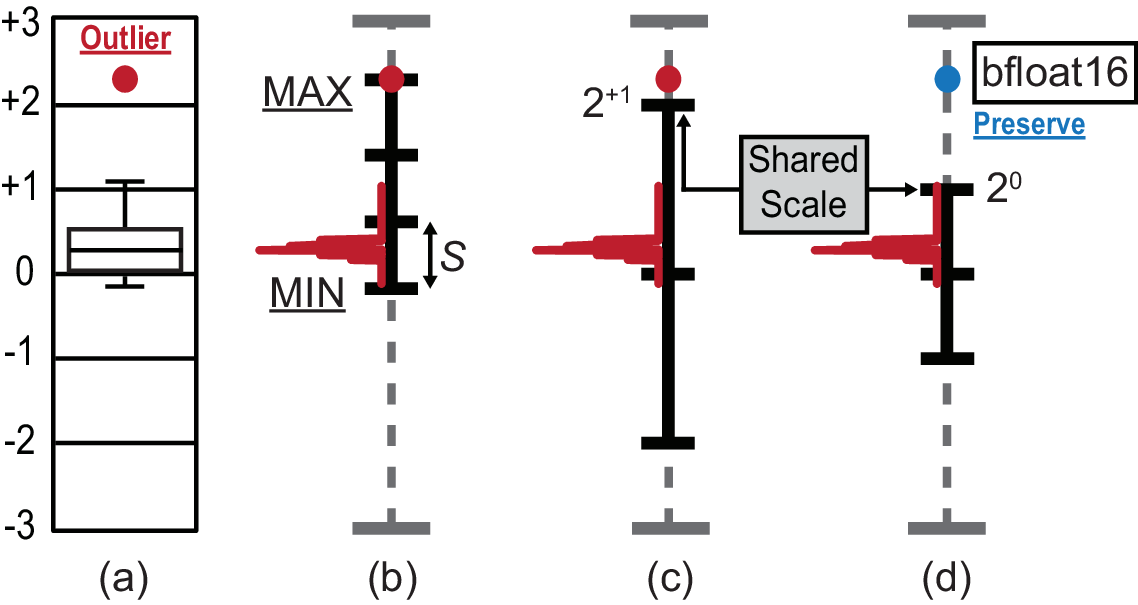}\vspace{-2mm}
  \centering
  \caption{Comparison between different data formats on quantizing the original data of 128 elements extracted from the 2nd decoder block in Llama2-7B. (a) Original data, (b) 2-bit MinMax, (c) MXINT2, and (d) MX-OPAL2.}
  %Compared quantization and MXINT. (a) is input distribution in \textit{self\_attn.o\_proj} on the 2nd decoder block of Llama-7b (b) is MinMax quantization of (a), (c) convert MXINT of (a), (d) convert Outlier-Preserved MXINT of (a)}
    \label{compared_quant} \vspace{-5mm}
\end{figure}

\subsection{Microscaling Data Format}
Recently, microscaling (MX) data formats~\cite{microscaling} have been presented showing their effectiveness on both inference and training of various deep learning benchmarks.
%is a new way of representing data types for training and inference in deep learning systems. 
The MX format packs `\textit{k}' data into a block and shares one exponent per block (i.e., \textit{shared scale}; 2$^s$), and normalizes each element in the block with the shared scale.
Each element can be represented by either a low bit-width floating-point (e.g., MXFP6) or integer number (e.g., MXINT8).
The integer version of an MX format (MXINT) is also known as the block floating point~\cite{dbps}.
Converting FP numbers to MXINT has two benefits: (i) \textbf{hardware efficiency}---reduced bit-width and INT-based arithmetic maximize computing and storage efficiency, 
and (ii) \textbf{simplified conversion}---unlike the conventional quantizer, which requires FP divisions by the scale factor, MXINT can scale each element by simple shift operations (suited at dynamic quantization). 

Fig.~\ref{BFP_format} shows an example of converting bfloat16 values to MXINT4. 
First, we find the shared scale which is the maximum exponent within a block (in this example, $k=6$). 
For instance, the maximum exponent is 130 in Fig.~\ref{BFP_format}(a), which becomes the shared scale. 
Then, mantissa bits are shifted right by the difference between the shared scale and its own exponent (Fig.~\ref{BFP_format}(b)).
Due to the shift operation, there could be underflow that results in the quantization error (crossed out by the red line).
% In Fig. \ref{BFP_format}(b), For the MXINT conversion, it has been shifted to the right by the difference between 130 and the exponent, and we can see that most values are now zero (\textit{red line}). 
Thus, it could be problematic if there is few significantly large outliers in the block, which makes other non-outliers zero.
As observed by the prior work and our analysis, there are such outliers in activations for LLMs.
%almost all the existing values will be changed to 0 when converted to MXINT format, which is fatal in LLM where there is an outlier in the activation. 

%converts \textit{k} original data into \textit{k} low-bit data and a shared scale \textbf{X} for the scalar element \textit{k}. The conversion process of MXINT consists of a search for the maximum exponent in the existing values and a shift to normalize the remaining values to match the found maximum exponent, as depicted in Fig. \ref{compared_quant} (c). Converting the MX format has two benefits: (1) \textbf{Hardware efficiency}---Reduced bit width to maximize computing and storage efficiency and diminish data movement. (2) \textbf{Simplified conversion data format}---Unlike traditional quantization for representing low bit, which requires division by a scale factor, MX can be simply represented via shift operation. 

%However, as shown in Figure 2, MXINT has the disadvantage of not adequately representing data where outliers are present, which breaks performance in LLM models. To address these issues, we propose an \textbf{\underline{O}}utlier-\textbf{\underline{P}}reserved \textbf{\underline{A}}daptive microscaling quantization for \textbf{\underline{L}}LM (OPAL),which does not degrade the performance of LLM by storing some of the outliers. And we also propose an OPAL-optimized accelerator for LLM's generation task.

\begin{figure}[t]
  \includegraphics[width=0.9\linewidth]{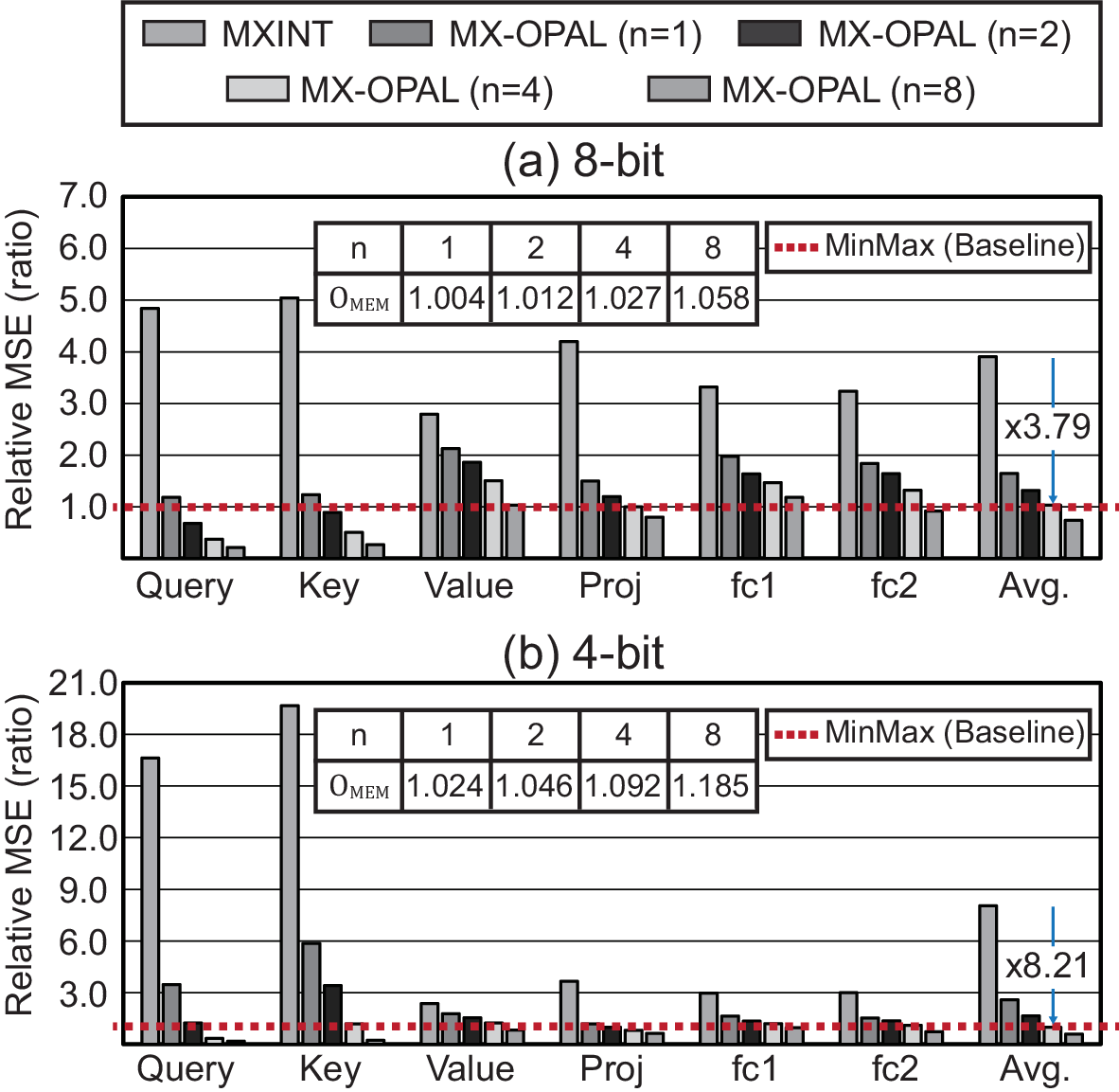}\vspace{-1mm}
  \centering
  \caption{Comparison of the impact of preserving varying number of outliers (\textit{n}) on the quantization noise with MX formats at the 20th decoder block in Llama2-7B. 
  The block size \textit{k} is set to 128, and (a) `sign + mantissa bits' = 8 ($b=8$), (b) `sign + mantissa bits' = 4 ($b=4$).}
    \label{fig:compared_outlier}\vspace{-5mm}
\end{figure}

\section{Proposed MX-OPAL Data Format} 
\subsection{Outlier-Preserved Microscaling Data Format}
To address this challenge, we propose an outlier-preserved microscaling format, named MX-OPAL, that maintains the LLM accuracy by keeping significantly large outliers in bfloat16. 
Note that we apply the MX-OPAL format for activations while using  OWQ~\cite{owq} for weights.
They are a good match since OWQ also maintains the weights in FP format at channels where activation outliers occur, while the remaining weights are stored in INT3 (or INT4).
An example of converting bfloat16 to 4-bit MX-OPAL is shown in Fig.~\ref{BFP_format}(c).
In MX-OPAL, we preserve the top-$n$ outliers where $n=1$ in the example.
After preserving the outlier(s), we extract the ($n$+1)\textsuperscript{th} highest exponent value as a shared scale.
In the proposed MX-OPAL format, we set a tensor-wise global shared scale (118 in Fig.~\ref{BFP_format}(c)) and store a 4-bit block-wise offset for every $k$ elements (10 in Fig.~\ref{BFP_format}(c)) to minimize the data size of shared scales.
Compared to MXINT4, we can reduce the number of underflows by using the proposed MX-OPAL4 as shown in Fig.~\ref{BFP_format}(c).
% \textbf{\underline{O}}utlier-\textbf{\underline{P}}reserved \textbf{\underline{A}}daptive microscaling quantization for \textbf{\underline{L}}LM Microscaling integer (OPAL-MX),which does not degrade the performance of LLM by preserving some of the outliers. And we also propose an OPAL-MX optimized accelerator for LLM's generation task.

Fig.~\ref{compared_quant} compares the quality of three different quantizers, i.e., 2-bit MinMax~\cite{zeroquant}, MXINT2~\cite{microscaling}, and MX-OPAL2, on actual data extracted from the input to the projection layer (\textit{self\_attn.o\_proj}) in the 2nd decoder block of Llama2-7B.
We extracted 128 elements from 4,096 elements in the input vector that makes a block (thus, 32 blocks are formed).
The data distribution of the first 128 elements is provided in Fig.~\ref{compared_quant}(a).
There is an outlier that is away from the other data.
The MinMax quantizer sets the outlier as the maximum value and finds the other end as the minimum.
Then, it divides each value by the scale factor $S=(Max-Min)/(2^b-1)$, where `$b$' is the bit-width, and rounds the result to the nearest neighbor.
The MXINT format finds the outlier and sets its exponent as the shared scale.
In Fig.~\ref{compared_quant}(c), the quantization level right below the outlier becomes this shared scale.
After setting the shared scale, which is the power of 2, the elements are quantized by the given `sign+mantissa' bits (2-bit in this example).
A large portion of elements then falls in the same quantization bin, resulting in a large quantization error. 
With the proposed MX-OPAL format, we keep the outlier in bfloat16 and finds the second largest exponent within the block (Fig.~\ref{compared_quant}(d)).
Effectively, the shared scale moves closer to the mean of the distribution and the step size becomes smaller (lowering the quantization error), since the smaller exponent has been selected as the scale factor.

% (a) is part of the input to the projection layer (self\_attn.o\_proj) of the 2nd decoder block of Llama-7b. 
% For such outliers, MXINT finds shared scale based on the maximum value in , which can sometimes differ massively from the original value, leading to performance degradation. Therefore, we propose to preserve some of the outliers to improve MXINT's generalizability. 
% When searching for the maximum of an exponent to fix the shared scale, we need to find not only the maximum but also the \textit{(n+1)th} maximum to preserved outlier. Then, \textit{nth} maximum value is stored as original \textit{bfloat16}, select the \textit{(n+1)th} value as shared exponent. For example, Fig. \ref{BFP_format} (c) shows OPAL-MX format \textit{k} = 6, \textit{n} = 1. To determine the shared scale, we found the maximum exponent value without preserving the outlier. In this example, the shared scale becomes 128, and as a result, the value that changes to 0 is significantly reduced.

\subsection{Impact of Preserving Activation Outliers}

Preserving several outliers allows us to push majority non-outliers to a low bit-width, e.g., 3-bit or 5-bit, while the outliers are kept in 16-bit and require bfloat16 computation in MX-OPAL. 
Thus, we have to make sure the memory overhead of storing those outliers is not critical.
The memory overhead ($O_{MEM}$) of the MX-OPAL format relative to the original MXINT (or MinMax) format can be formulated as follows:
% More precisely, The relative memory overhead obtained using MX-OPAL are as follows.
\begin{equation}\label{eq:mem_overhead}
    O_{MEM}=\frac{MEM_{OPAL}}{MEM_{MXINT}}=\frac{(k-n)\cdot b + 16\cdot n + 4}{k\cdot b + 8},
\end{equation}
where \textit{k} is \# of elements in a block, \textit{n} is \# of preserved outliers, \textit{b} is the bit-width of non-outliers.
The constant 4 is added in the numerator since we use 4-bit to store one block-wise shared scale per block.
According to Eq.~(\ref{eq:mem_overhead}), the memory overhead of storing tensors in MX-OPAL format becomes negligible when the block size `$k$' is large enough.
% shows efficiency when the block size \textit{k} is large or the length of the presentation bit \textit{b} is higher. 
For instance, only 2.7\% of additional memory space is needed when \textit{k} = 128, \textit{n} = 4, and \textit{b} = 8. 

In Fig. \ref{fig:compared_outlier}, we examined the impact of preserving outliers by computing the mean squared error (MSE) between the original bfloat16 inputs and their MX-OPAL quantized values at various layer outputs within the 20\textsuperscript{th} decoder block.
To show the relative quantization error compared to the conventional MinMax quantization, we normalized MSEs of MX-OPAL at varying numbers of outliers (i.e., $n$ = 1, 2, 4 and 8) to the MSE of the MinMax quantizer.
In addition, the relative MSEs of MXINT formats are also provided to show the effectiveness of preserving several outliers in MX-OPAL (3.79$\times$ and 8.21$\times$ lower error on average than MXINT for $b=8$ and $b=4$, respectively).
The quantization error becomes similar to the baseline (i.e., MinMax), or sometimes below the bar, when four outliers among 128 elements are preserved.
Thus, throughout the paper, we set the number of preserved outliers `$n$' to four.
When $n=4$, the memory overhead compared to the MinMax or MXINT format becomes 2.7\% and 9.2\% when $b=8$ and $b=4$, respectively.
% The y-axis shows whether it is greater by a factor of x than the MSE of the MinMax quantization. 
% MX has the disadvantage of not appropriately expressing existing values, but MX-OPAL only removes four outliers, which can be seen as an MSE similar to MinMax. So, we choose \# of preserving outlier is 4.

\section{Hardware Architecture of OPAL}\label{sec:OPAL_arch}

\subsection{OPAL Computation Flow}\label{sec:comp_flow}

\begin{figure}[t]
  \includegraphics[width=0.8\linewidth]{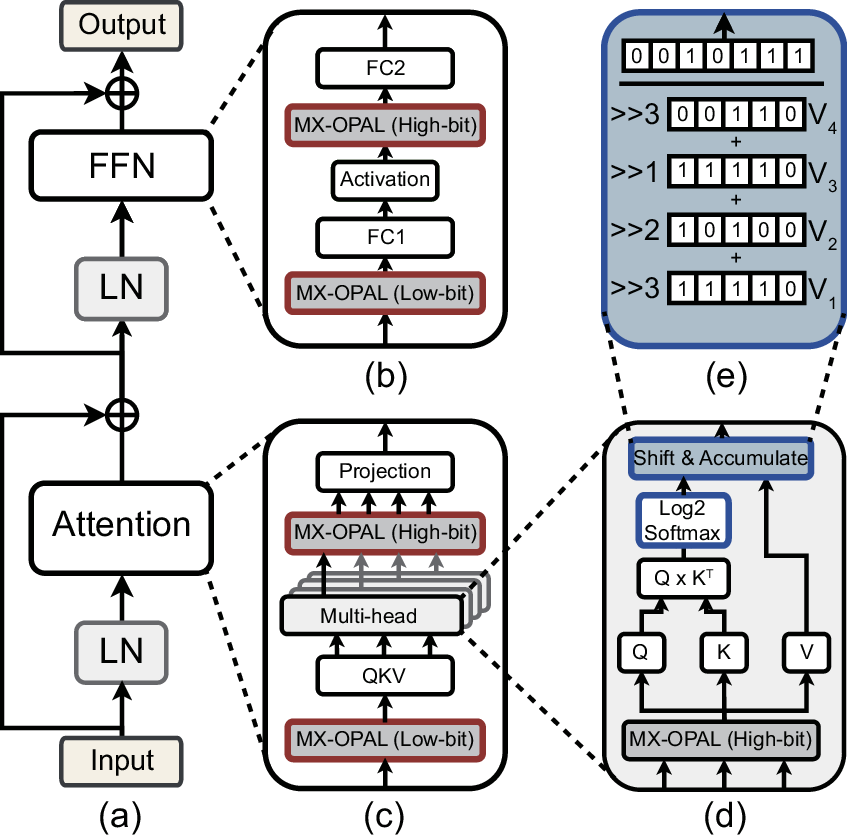}\vspace{-2mm}
  \centering
  \caption{Overview of OPAL computation flow: (a) one decoder block, (b) a feed forward network (i.e., two FC layers), (c) an attention layer, (d) a multi-head attention layer, and (e) shift-based `$Attn\cdot V$' owing to log2-based softmax.}
    \label{fig:transformer}\vspace{-5mm}
\end{figure}

Fig.~\ref{fig:transformer} shows the OPAL computation flow that is implemented as a hardware accelerator described in Section~\ref{sec:microarch}.
Fig.~\ref{fig:transformer}(a) shows one decoder block which is repeated 32/80 times to form Llama2-7B/70B.
There are two main computing layers which require high number of parameters: a feedforward network (FFN in Fig.~\ref{fig:transformer}(b)), and an attention layer (Fig.~\ref{fig:transformer}(c)).
Since activations are normalized by layer normalization (LN), the distribution of activations is limited to a specific range.
Thus, further reduced bit-width can be used, e.g., MX-OPAL3, to quantize activations after the LN layer.
Activations at other layers are kept in higher bit-width, e.g., MX-OPAL5, to maintain the accuracy at a similar level to 16-bit activations.
In the attention layer (Fig.~\ref{fig:transformer}(c-d)), multiple matrix-vector multiplications (MxV) are performed to compute $Q$, $K$, $V$, and the attention map ($Attn$).
Thanks to the proposed log2-based softmax unit, we simply perform shift-and-accumulate to compute the output $Z$ (Fig.~\ref{fig:transformer}(e)).

\subsection{Proposed Log\textsubscript{2}-based Softmax Unit}\label{sec:our_softmax}
In LLMs, softmax is one of the most hardware-unfriendly operations, that requires floating-point dividers, thus several algorithms have been proposed to approximate the softmax function~\cite{ivit,softermax}.
Unfortunately, the prior work require fine-tuning to compensate for the approximation error which is not suitable for LLMs.
% However, most approximation softmax require training, so there is a disadvantage that it is difficult to use in LLM task. 
Therefore, we modified the log2-based softmax method~\cite{fqvit}, an approach for PTQ, on optimizing LLM inference. 
The output of the log2-based softmax in the multi-head attention layer is defined as:
\begin{equation}\label{eq:log2_softmax}
    Attn_Q=clip(-\lceil log_2(softmax(\frac{Q \cdot K^T}{\sqrt{d_k}}))\rfloor, 0, 2^b-1),
\end{equation}
where $Q$ and $K$ are query and key, $d_k$ is the embedding dimension, and $b$ is the bit-width.
The $log_2(softmax(\cdot))$ always outputs a negative value since softmax result lies within (0, 1). 
% In addition, the existing softmax required a floating-point divider, but the log2 quantization softmax can only be replaced by an integrator to subtract the exponent. 
% There are two advantages of the log2-based softmax:
% (i) it is possible to express a wider value range compared to the uniform quantization.
% LLM's softmax results often have small values that can be expressed in fewer bits. 
Eq.~(\ref{eq:log2_softmax}) can be considered as the log2 quantization of the attention map ($Attn=softmax$($Q\cdot K^T/\sqrt{d_k}$)), since it can be computed by $2^{-Attn_Q}$.
With such log2 quantization, we can express a wider range of values in the attention map, and compute the ouput `$Z=Attn\cdot V$' with simple shift-and-accumulate operations. 
However, the previous work~\cite{fqvit} requires FP multipliers, FP dividers, and log2 unit to compute $log_2(softmax(\cdot))$.
% Since the result of log2 quantization is only a power of 2, the corresponding operation can be replaced by shift operation, the 
% same as subtracting the result of log2 quantization from the value of the existing v's exponent. 
In OPAL, we simplify the hardware of log2-based softmax by modifying the computation as follows:
\begin{equation}
\label{eq:our_softmax}
    \begin{split}
        \lceil log_2(\frac{e^{x_i}}{\Sigma e^{x_i}}) \rfloor &= \lceil log_2(e^{x_i}) - log_2(\sum e^{x_i}) \rfloor \\ 
        &= \lceil log_2(2^{E_i} \cdot 1.M_i) - log_2(2^{E_{\sum}} \cdot 1.M_{\sum}) \rfloor \\
        &= (E_i - E_{\sum}) + \lceil log_2(1.M_i) - log_2(1.M_{\sum}) \rfloor \\ 
        & = (E_i - E_{\sum}) + Sign(M_i-M_{\sum})\circ 1_{|M_i-M_{\sum}|\geq0.5}, 
    \end{split}
\end{equation}
where $E_i$ is the exponent of $e^{x_i}$, and $1.M_i$ is `1 + mantissa' of $e^{x_i}$, $E_{\sum}$ is the exponent of $\sum e^{x_i}$, and $1.M_{\sum}$ is `1 + mantissa' of $\sum e^{x_i}$.  
Since $1.M$ always exists between 1 and 2, $log_2(1.M)$ is a value between 0 and 1. 
As a result, we can compute the complex `$log_2(softmax(\cdot))$' function using INT subtractors for ($E_i - E_\Sigma$) and ($M_i - M_\Sigma$), and comparators.

\begin{figure}[t]
  \includegraphics[width=0.9\linewidth]{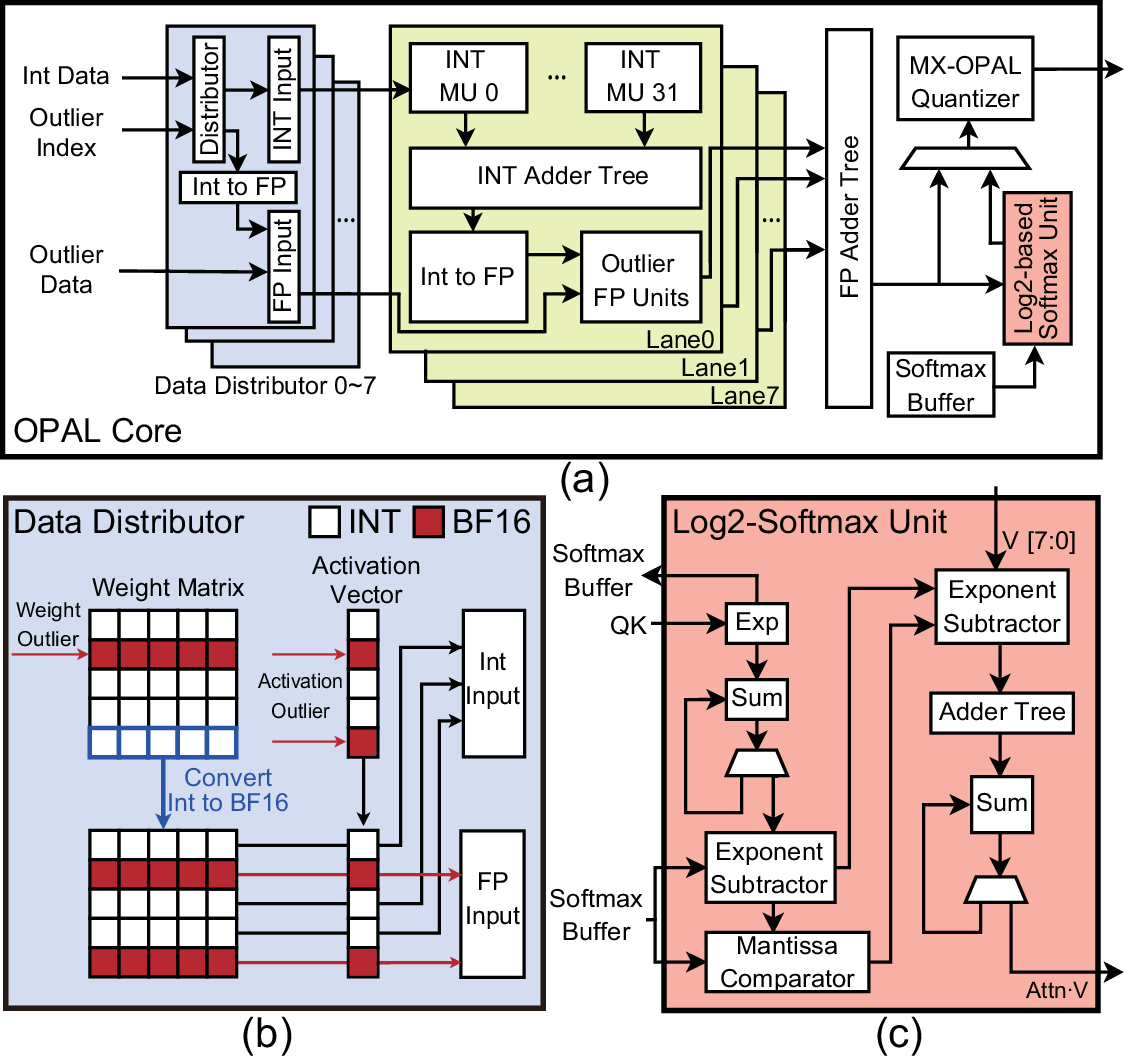}\vspace{-2mm}
  \caption{Microarchitecture of OPAL: (a) the main core of OPAL, (b) the data distributor for outliers and non-outliers, and (c) the proposed log2-based softmax unit.}
    \label{fig:block_diagram}\vspace{-5mm}
\end{figure}
\vspace{-2mm}
\subsection{OPAL Microarchitecture}\label{sec:microarch}

\subsubsection{Data Distributor}
The microarchitecture of the main core in OPAL is illustrated in Fig.~\ref{fig:block_diagram}(a).
% OPAL's accelerator is designed as shown in ~\ref{fig:block_diagram}(a). 
Each core consists of eight data distributors, eight MxV compute lanes, an FP adder tree, a log2-based softmax unit, and an MX-OPAL quantizer. 
To each core, 128 activations, either in low-bit INT (3 or 4-bit) or high-bit INT (5 or 7-bit) with one 4-bit shared scale, are provided as an input which includes four bfloat16 (BF16) outliers.
The data distributor attached in front of each compute lane routes non-outliers to INT multiply units (\texttt{INT MU}s) and outliers to \texttt{FP unit}s.
% Then, using the Data Distributor, do the following for separating INT and BF16 computing units. First, convert the value that operates on outliers to BF16. 
Since there exist more outliers in activations than weights ($\sim$3\% vs. $\sim$0.3\%), we convert some channels in the weight matrix that align with the activation outliers to BF16 from INT3/4 (highlighted in a blue box in Fig. ~\ref{fig:block_diagram}(b)).
% Second, distribute the values using the INT type as the INT operator and the FP type as the FP operator.
% As shown in Fig. ~\ref{fig:block_diagram} (b), If the index of the outlier is in the same position, change it to bfloat16 and move to the \texttt{FP Input}, and for the rest, move to the \texttt{INT Input}.
Most activations and weights are directed to \texttt{INT MU}s saving significant amount of power in OPAL.

\begin{figure}[t]
  \includegraphics[width=0.95\linewidth]{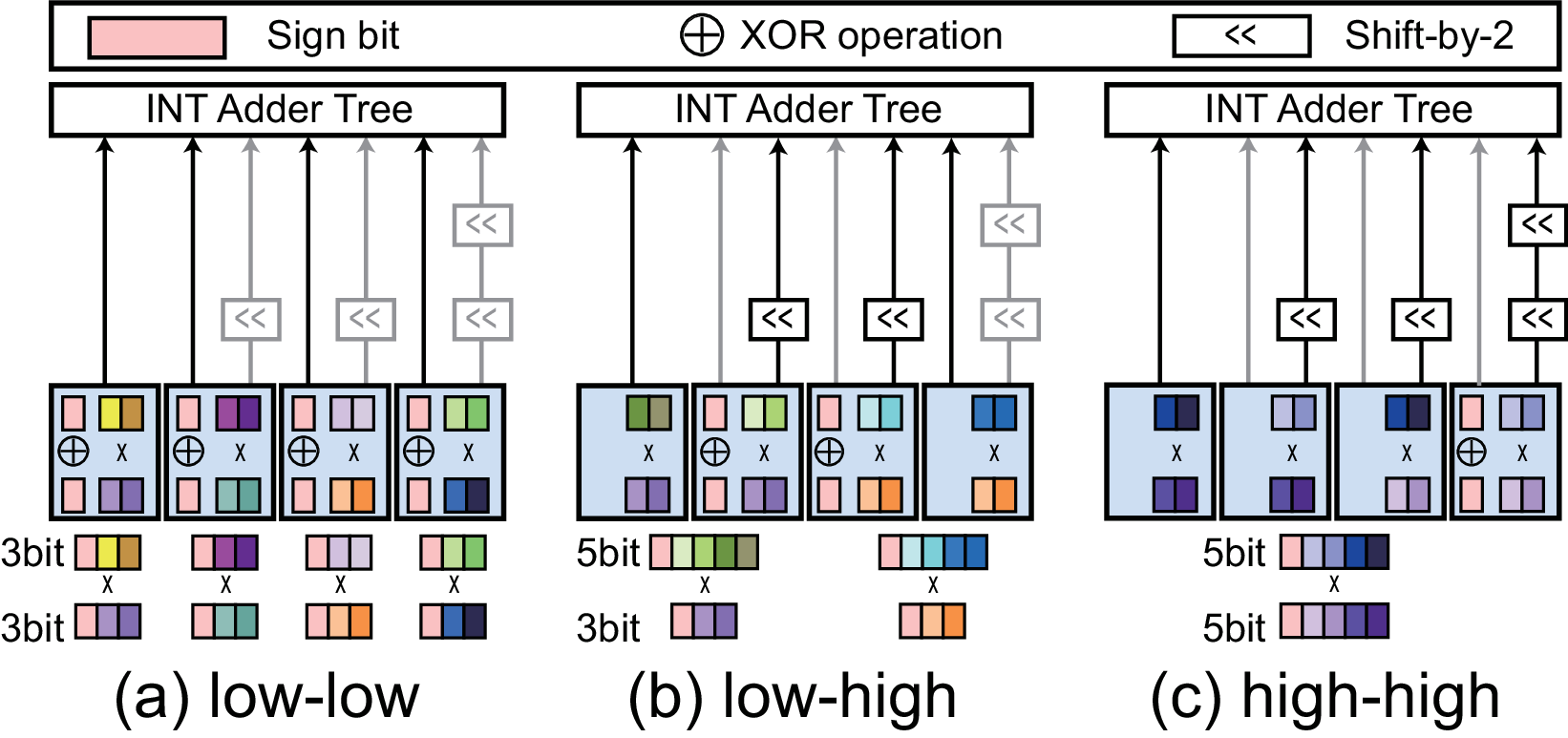}\vspace{-2mm}
  \centering
  \caption{Operation modes of an \texttt{INT MU}: (a) \texttt{low-low} mode, (b) \texttt{low-high} mode, and (c) \texttt{high-high} mode.}
    \label{fig:mu_chunk}\vspace{-4mm}
\end{figure}

\begin{table}[]
\caption{Perplexity ($\downarrow$; the lower, the better) of Llama2 and OPT tested on WikiText-2 at various quantization schemes}
%W is weight bit, A is activation bit. A4/7 means that the layer after LN is quantized to 4-bit and the other layer is quantized to 7-bit.}
\resizebox{\columnwidth}{!}{%
\begin{tabular}{|l|cc|cc|}
\hline
\textbf{Model}         & \multicolumn{1}{c|}{\textbf{Llama2-7B}} & \multicolumn{1}{c|}{\textbf{Llama2-13B}}                  & \multicolumn{1}{c|}{\textbf{OPT-6.7B}} & \multicolumn{1}{c|}{\textbf{OPT-13B}}                           \\ \hline\hline
bfloat16 (BF16) (Baseline)      & \multicolumn{1}{c|}{5.472}          & {4.884}          & \multicolumn{1}{c|}{10.854}         & {10.132}           \\ \hline
W4A16 (OWQ~\cite{owq})        & \multicolumn{1}{c|}{6.031}          &    {5.113}             & \multicolumn{1}{c|}{10.949}         &    {10.275}             \\ \hline  \hline
W4A7 (MinMax \cite{zeroquant})& \multicolumn{1}{c|}{6.693}          &       {5.267}          & \multicolumn{1}{c|}{10.903}         &    {10.179}             \\ \hline
W4A7 (MX-OPAL)& \multicolumn{1}{c|}{6.431}          &       {5.256}          & \multicolumn{1}{c|}{10.894}         &       {10.191}          \\ \hline
W4A4/7 (MinMax \cite{zeroquant})& \multicolumn{1}{c|}{6.546}          &  {5.472}               & \multicolumn{1}{c|}{12.071}         &     {11.586}            \\ \hline
W4A4/7 (MX-OPAL)   & \multicolumn{1}{c|}{6.492}          &  {5.302}               & \multicolumn{1}{c|}{10.977}         &      {10.312}           \\ \hline \hline
W3A16 (OWQ~\cite{owq})        & \multicolumn{1}{c|}{6.684}          &    {5.653}             & \multicolumn{1}{c|}{11.183}         &    {11.741}             \\ \hline  
W3A3/5 (MinMax \cite{zeroquant})   & \multicolumn{1}{c|}{32.747}          &  {10.835}               & \multicolumn{1}{c|}{28.742}         &      {95.791}           \\ \hline
W3A3/5 (MX-OPAL)   & \multicolumn{1}{c|}{7.400}          &  {6.168}               & \multicolumn{1}{c|}{14.343}         &      {16.225}           \\ \hline
\end{tabular}\label{tab:compared_model}
}
\end{table}

\subsubsection{Compute Lane}
% The OPAL has eight lanes for MxV operation. 
Each compute lane consists of 32 \texttt{INT MU}s, one of which contains four INT multipliers. 
Each lane also has four \texttt{FP Unit}s to handle outliers. 
% There are INT multiply units (MUs) for non-outliers and FP units for outliers.
Each \texttt{INT MU} can be reconfigured to operate at either 3- or 5-bit mode, or it can be designed to support 4- and 7-bit modes for more accurate LLM inference (Section~\ref{sec:OPAL_accuracy}) at a higher hardware cost.
% The inputs to the core consist of 128 low-bit data (3- or 4-bits) or 32 high-bit data (5- or 7-bits), 4 outlier data per operand. 
% We store activations in INT datatype, with one 4-bit shared exponent, except four BF16 outliers within a block.
In this work, all weights except outliers are stored in INT3 or INT4 (i.e., low-bit INT) based on OWQ~\cite{owq} method.
Thus, the \texttt{INT MU} in OPAL supports three modes, i.e., `\texttt{low-low}', `\texttt{low-high}', and `\texttt{high-high}' modes (Fig.~\ref{fig:mu_chunk}).
The \texttt{low-low} mode is set to perform MxV between non-outlier weights and low-bit activations (i.e., after LN layer).
The \texttt{low-high} mode is used for MxV between non-outlier weights and high-bit activations.
The \texttt{high-high} mode is used to perform MxV between high-bit activations, such as $Q\cdot K^T$.
Our \texttt{INT MU}s can flexibly support three modes, with the \texttt{low-low} mode providing 4$\times$ throughput over the \texttt{high-high} mode. 
The \texttt{INT MU} outputs are accumulated to a single output through an INT Adder Tree, and the result is converted to BF16 by incorporating the shared scale via an Int-to-FP unit.
Then, it is accumulated with BF16 results from outlier FP units. 

% As shown in Fig. \ref{fig:mu_chunk}, Int Multiplier consists of three modes. 
% There is a \texttt{low-low} mode that operates weight and low bit activation as a low bit, a \texttt{low-high} mode operates between low bit weight and high bit activation, and a \texttt{High2High} mode operates high bit activations to calculate the attention score.
% Our \texttt{INT MU chunks} can flexibly support three modes, with the \texttt{low-low} mode providing four times the throughput of the \texttt{high-high} mode. 
% As shown in Fig. \ref{fig:transformer}, We use all three modes to perform different activation quantizations in different layers to increase utilization 

\subsubsection{Softmax Unit and MX-OPAL Quantizer}
After MxV computations, the final output is generated by passing the results from eight lanes through the FP Adder Tree.
Then, outputs are selectively moved (for $Q\cdot K^T$) to log2-based softmax unit to transform them to attention map (Fig. ~\ref{fig:block_diagram}(c)).
By using our log2-based softmax, we can cut down 32.3\% of the area and 35.7\% of the power compared to the conventional softmax unit.
The 128 output values in BF16 are then grouped together and converted to the MX-OPAL format at the shift-based MX-OPAL quantizer to minimize the size of intermediate data.
% which is converted from \textit{bfloat16} (\texttt{BF16}) using MX-OPAL converter for computational efficiency and reduce memory footprint.
% \texttt{INT2FP} converters to handle outliers. 
% \texttt{INT MU Chunk}s supports vector multiplication of the integer type, where vector \texttt{INT MU} outputs are accumulated to a single output through a \texttt{INT Adder Tree}.
% Then, change the summed result to \textit{bfloat16} and combine it to the calculated result of the outlier.
% It is a dynamic quantizer for activations for the next layer which is a simple shift-based quantizer. 
Then, the output of the quantizer is connected to either LN hardware or SRAM that is outside the OPAL core.

\begin{table}[]
\caption{Performance of  Llama2 tested on language modeling with WikiText-2 (\texttt{Wiki}) and C4 (\texttt{C4}) (perplexity $\downarrow$), and on zero-shot question answering tasks with ARC\_Challenge (\texttt{ARC}) and Physical Interaction: Question Answering (\texttt{PIQA}) (accuracy $\uparrow$)}
\resizebox{0.9\columnwidth}{!}{%
\begin{tabular}{|l|c|c|c|c|}
\hline
\textbf{Task}                & \textbf{Wiki $\downarrow$} & \textbf{C4  $\downarrow$} & \textbf{ARC $\uparrow$} & \textbf{PIQA $\uparrow$} \\ \hline \hline
Llama2-7B OWQ W4A16  & 6.031       & 7.744     & 37.97        & 76.55      \\ \hline
Llama2-7B MX-OPAL W4A4/7  & 6.492       & 8.084     & 37.80        & 75.24      \\ \hline
Llama2-7B OWQ W3A16  & 6.684       & 8.645     & 37.46        & 75.63        \\ \hline
Llama2-7B MX-OPAL W3A3/5  & 7.400       & 9.588     & 35.49        & 73.29      \\ \hline \hline
Llama2-13B OWQ W4A16 & 5.113       & 7.051     & 42.75        & 77.74      \\ \hline
Llama2-13B MX-OPAL W4A4/7 & 5.302       & 7.313     & 42.41        & 77.69      \\ \hline
Llama2-13B OWQ W3A16 & 5.653       & 7.710     & 39.33        & 76.88      \\ \hline
Llama2-13B MX-OPAL W3A3/5 & 6.168       & 8.445     & 38.73        & 75.03      \\ \hline \hline
Llama2-70B OWQ W4A16 & 3.496       & 5.837     &  46.75           & 81.01          \\ \hline
Llama2-70B MX-OPAL W4A4/7 & 3.596       & 5.933     & 46.50        & 80.96      \\ \hline
Llama2-70B OWQ W3A16 & 3.997       & 6.227     &    47.61         &   80.46     \\ \hline
Llama2-70B MX-OPAL W3A3/5 & 4.348       & 6.573     &  45.81       & 79.11       \\ \hline
\end{tabular}\label{tab:llama_performance}\vspace{-5mm}
}
\end{table}

\begin{table}[]
\caption{Area and power breakdowns of one OPAL core}\vspace{-1mm}
\resizebox{0.85\columnwidth}{!}{%

\begin{tabular}{|l|r|r|}
\hline
\textbf{OPAL Core (\texttt{W4A4/7})}        & \multicolumn{1}{c|}{\textbf{Area (µm$^2$)} } & \multicolumn{1}{c|}{\textbf{Power (mW)}} \\ \hline \hline
Compute Lanes                 & 670,126.34 (72.11\%)                 & 229.65 (68.38\%)                     \\ \hline
Data distributors     & 139,713.48 (15.03\%)                 & 63.20 (18.82\%)                       \\ \hline
Log2-based Softmax Unit              & 76,330.92 (8.21\%)                   & 27.62 (8.22\%)                        \\ \hline
MX-OPAL Quantizer       & 34,670.88 (3.73\%)                   & 14.11 (4.20\%)                       \\ \hline
FP Adder Tree        & 8,470.80 (0.91\%)                     & 1.28 (0.38\%)                        \\ \hline \hline
\textbf{Total}                & \textbf{929,312.41}                          & \textbf{335.85}                               \\  \hline
\end{tabular}}
\label{tab:hardware_performance}\vspace{-4mm}
\end{table}

\section{Experimental Results}\label{sec:exp_results}
\subsection{Accuracy Analysis of MX-OPAL}\label{sec:OPAL_accuracy}
We modified block floating point (BFP) implementation in QPyTorch~\cite{qpytorch+} to evaluate our MX-OPAL format on inferencing LLMs at various language tasks. 
We have quantized activations prior to all MxV operations, as shown in Fig. \ref{fig:transformer}.
Quantizing to MX-OPAL on activations along with low-bit OWQ on weights allows mostly integer operations, reducing the latency of the LLM inference. 
Only channels where outliers happen at either activations or weights are handled by BF16 units.
Due to the high storage requirement of weights in LLM (making LLM memory-bounded), we perform aggressive quantization using OWQ~\cite{owq} to quantize weights to 4-bit (or 3-bit) with only 0.25\% (or 0.33\%) of BF16 outliers. 

In Table~\ref{tab:compared_model}, we compared the performance of Llama2 and OPT models on language modeling (WikiText-2) at various quantization schemes.
Activations after layernorm (LN) are pushed to a low bit-width (i.e., 4-bit or 3-bit) to maximize the efficiency, since the normalized distribution becomes more robust to quantization noise.
% better normalized than other activations, so fewer bits are used to increase computational efficiency. 
% We conjecture that the normalized activations are more robust to quantization noise, thus we quantized activations after LN to low-bit (\texttt{A4/7}, \texttt{A3/5}). 
Accordingly, we evaluate the activation quantization of \texttt{A4/7} and \texttt{A3/5} (i.e., low-bit/high-bit in Fig.~\ref{fig:transformer}).
All weights are quantized to the low-bit for all cases, i.e., 4-bit (\texttt{W4}) or 3-bit (\texttt{W3}).
With \texttt{W4A4/7}, MX-OPAL merely increases the perplexity (PPL) by 0.435 on average, while MinMax increases PPL by 1.083, when compared to the BF16 baseline.
The performance gap between the proposed MX-OPAL (0.616$\uparrow$ on Llama2) and the conventional MinMax quantization (3.822$\uparrow$ on Llama2) becomes much larger at \texttt{W3A3/5}.
Compared to OWQ, i.e., \texttt{W4A16} (or \texttt{W3A16}), MX-OPAL increases PPL by 0.325 (or 0.616) on Llama2 and 0.0325 (or 3.822) on OPT, respectively.
% 이 문장은 넣을지 말지? 넣는 게 좋을 듯?
Note that OPT performs worse even at BF16 and is more prone to aggressive quantization than Llama2 due to architectural difference.

% With more extreme quantization (\texttt{W3A3/5})
% For example, the difference between \texttt{W4A4} and \texttt{W4A4/7} MinMax quantization in the llama model is 2.01, while in MX-OPAL, it is a trivial difference of 0.054.
% When using 3-bit weight quantization \texttt{W3}, which uses extremely low bits to overcome memory bound, the performance gap is enormous between MinMax and MX-OPAL. 
% For example, MinMax quantization shows at least a $2\times$ increase over conventional perplexity, while MX-OPAL shows a $0.1\times$ increase for Llama and a $0.5\times$ increase for OPT.

Thus, we evaluated MX-OPAL in more detail using Llama2 on various language tasks with WikiText-2, C4, ARC\_Challenge, and PIQA datasets.
% In Tab 2., we conducted additional experiments on diverse tasks. 
% We compare our MX-OPAL and OWQ with low-bit weight quantization to address the memory bound.
Since the main focus of MX-OPAL is in activation quantization, we compared all benchmarks to OWQ which uses BF16 for activations, while keeping the weights in the same bit-width.
% perplexity 사용 벤치마크와 accuracy 사용 벤치마크에 대해서 W4A4/7 (및 W3A3/5) 사용시 정확도 차이를 평균값으로 리포트(7B, 13B, 70B 전체에 대해 평균 취해서)
When using \texttt{W4A4/7}, MX-OPAL only increases PPL by 0.241 (on Wiki and C4) and loses accuracy by 0.36\% (on ARC and PIQA), on average.
With more aggressive quantization, i.e., \texttt{W3A3/5}, MX-OPAL increases PPL by 0.601 (on Wiki and C4) and loses accuracy by 1.65\% (on ARC and PIQA), on average.
% In summary, the perplexity degradation is within 0.943, and the accuracy degradation within 2.34\% when using the proposed MX-OPAL format.

\vspace{-4mm}
\subsection{Hardware Efficiency of OPAL}
To evaluate the proposed OPAL hardware in detail, we implemented RTL of all building blocks shown in Fig. \ref{fig:block_diagram} excepts SRAMs. 
Then, we used Synopsys Design Compiler to estimate power consumption and area based on 65nm CMOS technology.
% To estimate the energy consumption of on-chip memory, including the leakage, CACTI~\cite{cacti} and OpenRAM~\cite{openRAM} are used.
To estimate the energy consumption of on-chip memory, including the leakage power, CACTI~\cite{cacti} was used.
% For power analysis, the energy consumption and cycle time to access the global buffer used CACTI\cite{cacti}, while softmax's local SRAM used openRAM\cite{openRAM}.
We assume an OPAL hardware with a 512KB global buffer for weights and activations, and a 2KB softmax buffer within the OPAL core. 
% 여기에 hardware power & energy consumption 된 결과를 넣어주면 좋을 것 같음.
%table 3은 OPAL 4/7의 area와 power breakdown이다. 76.9% area와 73.9%의 power가 메인 연산 Lane에 사용된다. (번역 예정-구자현) 
The eight lanes in each OPAL core are capable of computing $32\times 8=256$ MACs in the \texttt{high-high} mode.
The number of operations ramps up to 512 and 1,024 in the \texttt{low-high} and \texttt{low-low} modes, respectively. 
Table~\ref{tab:hardware_performance} reports the area and power breakdowns of one OPAL core that uses a \texttt{W4A4/7} format.
As expected, most of the power and area (72\% and 68\%, respectively) is consumed by eight compute lanes.
% performance 비교하는 걸 여기 추가하면 좋을 듯?
%1 token input 기준으로 시뮬레이션을 진행했다. 각각의 core는 8개를 사용했고, 이는 일반적으로 사용되는 multihaed의 개수가 8의 배수이기 때문이다. OPAL 3/5, OPAL 5/7코어와 일반적으로 사용되는 bfloat 16 코어를 비교하였고, throughput을 동일하게 설정하였다. Bfloat 16 코어 대비 area는 32.20% (3/5), 42%(4/7)의 area를 사용한다. LlaMa-2-70B 모델에서는 49%, 66%의 energy로 연산이 가능하다. (번역 예정-구자현)

\begin{figure}[t]
  \includegraphics[width=\linewidth]{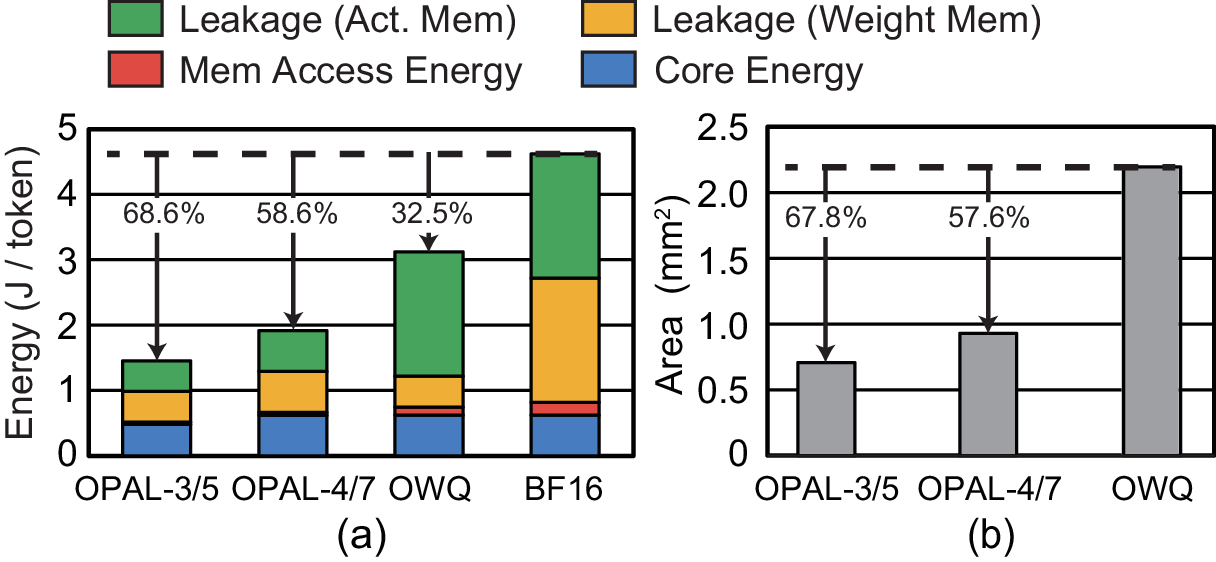}\vspace{-2mm}
  \centering
  \caption{Comparison on (a) the energy consumption of generating one token with Llama2-70B and (b) the area between OPAL and other baselines (OWQ~\cite{owq} and BF16).}
    \label{fig:energy_result}\vspace{-4mm}
\end{figure}

To evaluate the area and energy efficiency of the OPAL hardware, we have set two baselines, i.e., an BF16-based and an OWQ-based accelerator.
Since the inference of an LLM-based generation task is latency critical, a large portion of on-chip area is occupied by the global buffer.
The main challenge in deploying a large on-chip buffer lies not only in the large memory footprint but also in the high leakage power.
Fig.~\ref{fig:energy_result} compares the energy consumption of generating one token with Llama2-70B and the area of two OPAL hardware variants (\texttt{W3A3/5} and \texttt{W4A4/7}) to two baselines.
The latency of generating one token was 1.98 sec for Llama2-70B on OPAL.
The OWQ saves 32.5\% of the energy consumption on average by shrinking down the weight buffer while keeping intermediate activation data in BF16 (+ all computations are done in BF16 units).
By utilizing the proposed MX-OPAL format and its dedicated hardware, we can save the energy consumption by 53.5\%/68.6\% with \texttt{W3A3/5} and 38.6\%/58.6\% with \texttt{W4A4/7} on average compared to OWQ/BF16, respectively.
The OPAL hardware saves the energy by reducing the required size of on-chip buffer (effectively, reducing the leakage energy), and by simplifying the design of the main core.
%(e.g., INT MUs, log2-based softmax units, and shift-based quantizers).

% \usepackage{graphicx}

\vspace{-4mm}
\section{Conclusion}
In this paper, we proposed OPAL, an energy-efficient hardware accelerator for generative LLMs.
%hardware-software co-design accelerator for efficient LLM generation tasks.
To achieve high energy efficiency, we focused on quantizing activations to low bit-width INT format with a shared scale, i.e., MX-OPAL format, while preserving several outliers for every 128 elements.
% we first quantized the LLM weights to solve the memory bound, and then we quantized activation to increase the computational efficiency through MX-OPAL, a quantization method robust to the activation's outlier.
By using the proposed MX-OPAL format, 96.9\% of computations are done in INT multipliers significantly saving the computing energy.
In addition, we could reduce the required size of on-chip activation buffers owing to the lower activation bit-width significantly saving the leakage energy in the global buffer.
Overall, we were able to reduce the total energy of LLM inference on generation tasks by up to 46.5\% using the \texttt{W3A3/5} MX-OPAL format compared to the weight-only quantization method (OWQ), with negligible accuracy loss.
% We also approximated the softmax as a log2-based softmax for executing hardware-friendly operation and used different precision for each layer to maximize throughput of the OPAL core.
% Compared to the traditional FP-based architecture, the OPAL architecture synthesized using the ASIC design flow shows area and power efficiency with a slight performance penalty.
%We believe that our theoretical insights will expand the understanding of weight quantization, inspiring future research and promoting the widespread adoption of LLMs.
\vspace{-4mm}
\section*{Acknowledgment}
This work was partially supported by the Institute of Information \& Communications Technology Planning \& Evaluation (IITP) grant (RS-2023-00229849), and the National Research Foundation of Korea (NRF) grant (NRF-2023R1A2C2006290) funded by the Korean government (MSIT).
\vspace{-2mm}

\bibliographystyle{ACM-Reference-Format}
\bibliography{sample-base}


\end{document}